\documentclass[journal]{IEEEtai}

\pdfoutput=1

\usepackage[colorlinks,urlcolor=blue,linkcolor=blue,citecolor=blue]{hyperref}

\usepackage{color,array}

\usepackage{graphicx}
\usepackage{cite}
\usepackage{amsfonts}
\usepackage{amsthm}
\usepackage{amsmath}
\usepackage{algorithmic}
\usepackage{listings}
\usepackage{orcidlink}

\theoremstyle{definition}
\newtheorem{defn}{Definition}

\newcommand{\pre}[1]{\mathit{pre}(#1)}  %action preconditions
\newcommand{\eff}[1]{\mathit{eff}(#1)}  %action effects

\newcommand{\stCond}[1]{\mathit{startCond}(#1)}  
\newcommand{\enCond}[1]{\mathit{endCond}(#1)}  

\newcommand{\durCond}[1]{\mathit{durCond}(#1)}  
\newcommand{\duration}[1]{\mathit{dur}(#1)}  
\newcommand{\stEff}[1]{\mathit{startEff}(#1)}  
\newcommand{\enEff}[1]{\mathit{endEff}(#1)}  
\newcommand{\ctEff}[1]{\mathit{contEff}(#1)}  
\newcommand{\invCond}[1]{\mathit{inv}(#1)} 
\newcommand{\stAct}[1]{#1_\vdash} 
\newcommand{\enAct}[1]{#1_\dashv} 

\begin{document}

\title{Efficient Temporal Piecewise-Linear Numeric Planning with Lazy Consistency Checking}

\author{Josef~Bajada \orcidlink{0000-0002-8274-6177},
	Maria~Fox \orcidlink{0000-0002-1213-9283},
	and~Derek~Long \orcidlink{0000-0002-5720-9265}% <-this % stops a space
	\thanks{Josef Bajada is a Lecturer at the Department of Artificial Intelligence, University of Malta. e-mail: josef.bajada@um.edu.mt}% <-this % stops a space
	\thanks{Maria Fox is a Principal Researcher in Environmental AI at British Antarctic Survey.}
	\thanks{Derek Long is a Scientific Advisor at Schlumberger Cambridge Research.}% <-this % stops a space
}

\markboth{Journal of IEEE Transactions on Artificial Intelligence, Vol. 00, No. 0, Month 2021}
{Josef Bajada \MakeLowercase{\textit{et al.}}: Efficient Temporal Piecewise-Linear Numeric Planning with Lazy Consistency Checking}

\maketitle

\IEEEpubid{\begin{minipage}{\textwidth}\ \\[12pt] \centering
		\copyright2021 IEEE. Personal use of this material is permitted. Permission
	from IEEE must be obtained for all other uses, in any current or future
	media, including reprinting/republishing this material for advertising or
	promotional purposes, creating new collective works, for resale or
	redistribution to servers or lists, or reuse of any copyrighted
	component of this work in other works.\end{minipage}}

\begin{abstract}
	Temporal planning often involves numeric effects that are directly proportional to their action's duration. These include continuous effects, where a numeric variable is subjected to a rate of change while the action is being executed, and discrete duration-dependent effects, where the variable is updated instantaneously but the magnitude of such change is computed from the action's duration. When these effects are linear, state--of--the--art temporal planners often make use of Linear Programming to ensure that these numeric updates are consistent with the chosen start times and durations of the plan's actions. This is typically done for each evaluated state as part of the search process. This exhaustive approach is not scalable to solve real-world problems that require long plans, because the linear program's size becomes larger and slower to solve. In this work we propose techniques that minimise this overhead by computing these checks more selectively and formulating linear programs that have a smaller footprint. The effectiveness of these techniques is demonstrated on domains that use a mix of discrete and continuous effects, which is typical of real-world planning problems. The resultant planner also outperforms most state-of-the-art temporal-numeric and hybrid planners, in terms of both coverage and scalability. 
\end{abstract}

\begin{IEEEImpStatement}
While current temporal planners perform well on benchmark domains that feature continuous or duration--dependent numeric effects, real-world problems are often more challenging to solve. This is mostly due to the complex interactions between the temporal and numeric trajectory dynamics, together with the size of the plans that such planning problems typically require. The techniques proposed in this work take advantage of the structure of such domains to find a plan more efficiently. Empirical testing shows a significant improvement on the time spent on solving linear programs during the search for a plan, with an overall improvement on the total planning time for problems that exhibit such characteristics. Furthermore, the proposed planner outperforms all other state-of-the-art temporal-numeric planners in terms of both coverage and scalability. This is a significant achievement that enables A.I. Planning technology to solve a wider range of problems for industry applications. 
\end{IEEEImpStatement}


\section{Introduction}

\IEEEpubidadjcol

\IEEEPARstart{T}{emporal} planning is particularly challenging because the solver is not only concerned with discrete states, but also with temporal constraints that the plan's schedule must respect \cite{Cushing2007}. Real-world problems often pose rich numeric dynamics, such as continuous effects, where a numeric state variable is subjected to a rate of change while the action is being executed. These characteristics introduce complex interactions between the trajectory of numeric state variables and the temporal consistency of the plan's schedule. For example, if a plane spends less time refuelling than necessary, it will not have enough fuel to complete its journey. On the other hand, the longer it spends refuelling, the more the flight departure will be delayed, possibly missing its runway slot. We refer to numeric state variables whose value changes over the progression of the plan as \textit{numeric fluents} \cite{Fox2003,Hoffmann2003,Shin2005,Gerevini2008,Eyerich2009}. In the above example the fuel level is a numeric fluent.

\IEEEpubidadjcol

The prevailing approaches used by PDDL2.1~\cite{Fox2003} compliant planners that support such temporal--numeric interaction can be broadly categorised into two. The first category of temporal-numeric planners makes use of a \textit{Linear Program}~(LP) to model the interactions between temporal and numeric constraints. The second category of planners are more ambitious and try to support the full expressiveness permitted by PDDL2.1~\cite{Fox2003} and PDDL+~\cite{Fox2006a}, even when non-linear numeric dynamics are at play. These systems use either a discretise and validate approach, where the trajectory of numeric fluents is sampled at a chosen frequency, or Satisfiability Modulo Theories~(SMT) to support a suite of non-linear polynomial dynamics. In both categories, these systems are typically \textit{satisficing} planners, and do not guarantee solution optimality. 

The LP-based planners typically impose linearity restrictions, which are often an acceptable trade-off in return for higher performance and scalability. However, when used to solve larger problems that necessitate longer plans, the overheads of computing the LP start to accumulate~\cite{Denenberg2019}, impairing the planner from producing a solution within an acceptable time frame. The focus of this work is to improve the efficiency of LP-based planners so that they have a greater potential to be used successfully in real-world applications. More specifically, this work is relevant for PDDL2.1~Level~4~\cite{Fox2003} forward-search planners that make use of LPs to handle numeric effects that depend on an action's duration and check for temporal-numeric consistency.

In this work, we propose a more selective approach to decide when the LP for the current state needs to be solved, without introducing consistency or completeness trade-offs. We analyse the action that leads to the current state to determine whether running the LP provides any additional information to what is already known from the predecessor state. This technique proves to be effective in domains that have a mix of durative and instantaneous actions, especially in problem instances where effects that depend on their action's duration are localised to small parts of the plan, and thus, solving the LP for other states is unnecessary. This approach can also be used when evaluating goals whose terms include numeric fluents that are subject to such temporal interactions.
   
The main contributions of this work are as follows:

\begin{enumerate}
	\item A mechanism that selectively runs the LP solver only when the applied action imposes new constraints that can invalidate the plan's temporal-numeric consistency.
	\item An algorithm to minimise the number of LP runs needed to evaluate numeric goals. 
	\item An optimisation of the LP encoding, such that it is smaller and faster to solve.
\end{enumerate} 

%%%%%%%%%%%%%%%%%%%%%%%%%%%%%%%%%%%%%%%%%%%%%%%%%%%%%
\section{Background}\label{sec:Background}

A temporal planning problem that follows PDDL2.1 semantics \cite{Fox2003} is defined as $\Pi = \langle \rho, \vartheta, O_{inst}, O_{dur}, s_0, g \rangle$, where:

\begin{itemize} 
	\item $\rho$ is the set of atomic propositional facts,
	\item $\vartheta$ is the set of real-valued numeric fluents,
	\item $O_{inst}$ is the set of grounded instantaneous actions,
	\item $O_{dur}$ is the set of grounded durative actions,
	\item $s_0$ is the initial state, and
	\item $g$ is the goal condition. 
\end{itemize}

\begin{figure}[h]
	\centering
	\includegraphics[width=1.0\linewidth]{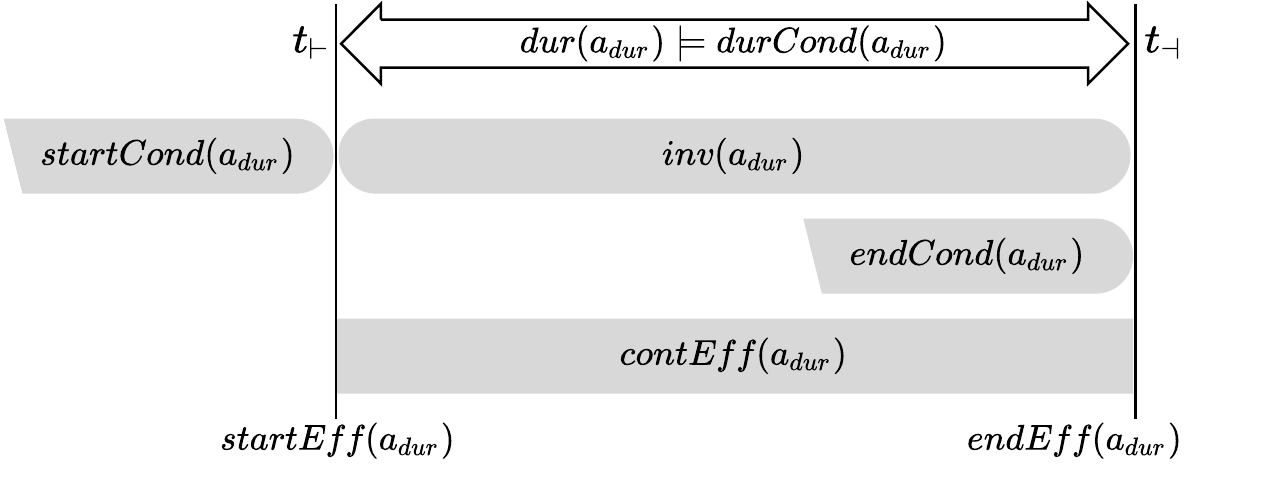}
	\caption[Durative Action]{The components of a PDDL2.1 durative action.} 
	\label{fig:DurativeAction}
\end{figure}

An \textit{instantaneous action}, $a_{inst} \in O_{inst}$, can have a precondition, $\pre{a_{inst}}$ and a set of discrete effects, $\eff{a_{inst}}$. Given the set of all possible states, $S$, for $\Pi$, for $a_{inst}$ to be applicable in $s$, the precondition must be satisfied by $s$, denoted $s \models \pre{a_{inst}}$. When the action is applied to a state $s$, a new state, $s'$, is obtained from applying its effects, $s' = \eff{a_{inst}}(s)$. A \textit{durative action}, $a_{dur} \in O_{dur}$, can have preconditions and effects at each of its endpoints together with temporal constraints, invariant conditions and continuous effects that hold throughout its execution, as shown in Figure~\ref{fig:DurativeAction}, where $t_{\vdash}$ and $t_{\dashv}$ correspond to the time points of the start and end endpoints of $a_{dur}$. 

The components of a durative action, $a_{dur} \in O_{dur}$, are defined as follows:

\begin{itemize}
	\item $\durCond{a_{dur}}$ is a conjunction over a set of temporal constraints on the action's duration, $\duration{a_{dur}}$, of the form $lb \leq \duration{a_{dur}} \leq ub$.
	\item $\stCond{a_{dur}}$ is the condition that needs to be true in some half-open interval $[t, t_{\vdash})$ where $t < t_{\vdash}$.
	\item $\enCond{a_{dur}}$ is the condition that needs to be true in some half-open 
	interval $[t, t_{\dashv})$ where $t < t_{\dashv}$.
	\item $\invCond{a_{dur}}$ is the invariant condition that needs to hold for every time instant $t$ in the open interval $(t_{\vdash}, t_{\dashv})$.
	\item $\stEff{a_{dur}}$ is the set of discrete effects that take place when the durative action starts at $t_{\vdash}$.
	\item $\enEff{a_{dur}}$ is the set of discrete effects that take place when the durative action ends at $t_{\dashv\textbf{}}$.
	\item $\ctEff{a_{dur}}$ is the set of continuous numeric effects applying over the closed interval $[t_{\vdash}, t_{\dashv}]$.
\end{itemize}

The conditions $pre(a)$, $startCond(a_{dur})$, $inv(a_{dur})$, and $endCond(a_{dur})$ are a conjunction of propositional logic formulas and numeric inequalities of the form $\langle e, \bowtie, k \rangle$, where $e$ is an arithmetic expression over a subset, $\vartheta_c \subseteq \vartheta$, of numeric fluents, $\bowtie\ \in \{ <, \leq, =, \geq, > \}$ is a comparison operator, and $k \in \mathbb{R}$ is a numeric constant. 

The discrete (instantaneous) effects can either be propositional effects making facts true or false, or numeric effects of the form $\langle v, \otimes, e \rangle$, where $v \in \vartheta$, referred to as the \textit{lvalue}, is the numeric fluent whose value is being updated, $\otimes \in \{ \mathbin{{:}{=}}, \mathbin{{+}{=}}, \mathbin{{-}{=}}, \mathbin{{*}{=}}, \mathbin{{/}{=}} \}$ is the assignment operator, and $e$ is an arithmetic expression, referred to as the \textit{rvalue} \cite{Fox2003}. 

The start and end endpoints of $a_{dur}$, denoted as $a_\vdash$ and $a_\dashv$ respectively \cite{Benton2012,Coles2014,Coles2016}, are themselves instantaneous actions, with $\pre{a_\vdash} = \stCond{a_{dur}}$, $\eff{a_\vdash} = \stEff{a_{dur}}$, $\pre{a_\dashv} = \enCond{a_{dur}}$, and $\eff{a_\dashv} = \enEff{a_{dur}}$. These are referred to as the \textit{snap actions} of durative action $a_{dur}$~\cite{Coles2008a, Gerevini2010, Coles2014,Bajada2015,Cashmore2018,Gigante2020}. Breaking down a durative action into its snap actions enables the planner to reason with concurrent actions and support \textit{required concurrency} \cite{Cushing2007} (when all possible solutions are concurrent), since it allows durative actions to overlap or enclose one another \cite{Coles2008a}. 

The set $O_s = O_{inst} \cup \{ a_\vdash, a_\dashv | a_{dur} \in O_{dur} \}$ consists of all discrete actions representing an instantaneous state transition. The time point at which a discrete action, $a \in O_s$, takes place in the plan is called a \textit{happening} \cite{Fox2003,Fox2004,Edelkamp2004,Gerevini2009,Coles2017,Cashmore2018}. 
The invariant condition, $\invCond{a_{dur}}$, of a durative action, $a_{dur}$, needs to hold during the open interval between its snap action's happenings. 

The continuous effects of a durative action are numeric expressions of the form $\langle v, e_\tau \rangle$, where $e_\tau$ defines the continuous change on a numeric fluent, $v \in \vartheta$, throughout the whole execution interval of the action. A durative action can run concurrently with other durative or instantaneous actions, which possibly also affect $v$ or $e_\tau$. We refer to the time interval $[t_i, t_{i+1}]$ enclosed by two consecutive happenings where $t_i < t_{i+1}$, as a \textit{temporal context} \cite{Bajada2016}. If $e_\tau$ is constant within a temporal context, then the continuous effect is linear for that temporal context \cite{Bajada2016}.

The start and end discrete effects of a durative action, $\stEff{a_{dur}}$ and $\enEff{a_{dur}}$ respectively, can also refer to the action's duration in their numeric expressions \cite{Fox2003}, making such effects \textit{duration-dependent}.

\begin{defn}
	A \textit{duration-dependent effect} on a numeric fluent $v$, is an effect of a durative action, $a_{dur}$, that modifies the value of $v$ with respect to the action's duration. This can be either:
	\begin{enumerate}
		\item a continuous effect, increasing or decreasing the value of $v$ gradually at a specific rate of change, or
		\item a discrete numeric effect on $v$ at the start or end of $a_{dur}$, with the total duration, $\duration{a_{dur}}$, being part of the effect's rvalue.
	\end{enumerate} 	
\end{defn}

A \textit{temporal plan} is a sequence of pairs, $\pi = (\langle t_1, a_1 \rangle, \langle t_2, a_2 \rangle, ... \langle t_n, a_n \rangle)$ \cite{Fox2003}, where $t_i > 0$ is the selected time point with respect to the start of the plan at which the action $a_i \in O_s$ is to be applied, and $n$ corresponds to the number of happenings in the plan. For an action $a \in O_s$, the state transition function for $\Pi$ is defined as $\gamma(s,a) = \eff{a}(s)$, given that $s \models \pre{a}$. 

Assuming all continuous effects are linear, given two time points $t_i$ and $t_{i+1}$ of two consecutive happenings, the value of $v$ at $t$, where $t_i \leq t \leq t_{i+1}$ is $v_i + \frac{dv}{dt} (t - t_i)$, with $v_i$ denoting the value of $v$ in $s_i$ (the state obtained from the happening at $t_i$). Since multiple continuous effects, $(\langle v,e_{\tau,1} \rangle, \langle v, e_{\tau,2} \rangle, ... \langle v, e_{\tau,m} \rangle)$, can be running concurrently on the same numeric fluent, $v$, the cumulative rate of change on $v$ is $\frac{dv}{dt} = \sum_{l=1}^{m} e_{\tau,l}$. Since all the continuous effects are linear, $\frac{dv}{dt}$ is constant throughout the time interval between $t_i$ and $t_{i+1}$ \cite{Bajada2016}. From this point onwards, we will be exclusively dealing with linear continuous effects. 

The temporal constraints of the plan, corresponding to the duration constraints, $\durCond{a_{dur}}$, of each durative action, $a_{dur}$, in the plan, are typically represented as a system of inequalities over its happenings. These take the form $lb \leq t_j - t_i \leq ub$, where $t_i$ and $t_j$ are the time points of two ordered happenings that need to happen at least $lb$ and at most $ub$ time units apart from each other. This system is referred to as a Simple Temporal Network (STN)~\cite{Dechter1991}. A problem that can be encoded as an STN is referred to as a Simple Temporal Problem~(STP)~\cite{Dechter1991}. 

The \textit{distance graph}~\cite{Dechter1991} of an STN is a weighted directed graph. Each node represents a discrete happening. Each edge is labelled with a positive weight if it defines an upper bound duration between two nodes, or a negative weight in the opposite direction if it represents a lower bound. An STN is inconsistent if its distance graph has at least one \textit{negative cycle}~\cite{Dechter1991}. Checking that an STN is consistent thus ensures that the plan's schedule is temporally consistent. This can be interleaved within the search process~\cite{Coles2009,Gerevini2010}. Significant work has also been done to optimise the representation and update operations of an STN~\cite{Muscettola1998,Planken2010,Planken2012,Lee2016} to perform efficient incremental updates and minimise its computational and memory footprint.

A \textit{temporal state} for a temporal planning problem $\Pi$ is defined as $s = \langle F, V, P, Q, C \rangle$ where:
\begin{itemize}
	\item $F$ is the set of atomic propositional facts from $\rho$ known to be true in $s$ (a closed world is assumed),
	\item $V$ is a set of bounds of the form $lb \leq v \leq ub$ on the possible values of the numeric fluents $v \in \vartheta$,
	\item $P$ is a partial plan, a list of instantaneous actions $(a_0, a_1, ..., a_n)$ needed to reach $s$ from $s_0$, where $a_i \in O_s$,
	\item $Q$ is the list of running durative actions, and
	\item $C$ is the STN for partial plan $P$, a set of temporal constraints of the form $lb \leq t_j - t_i \leq ub$. 
\end{itemize}

The constraints encoded in the STN are not enough when durative actions have  \textit{duration-dependent effects}. If such actions have a flexible duration that needs to be chosen by the solver, the value of such numeric fluents at any point in time depends on the chosen duration for the action. Furthermore, in the case of continuous effects, if the durative action is still running, the value depends on the time elapsed since the action started. Other actions that have numeric conditions on such numeric fluents are subject to more complex temporal constraints, since they might be restricted to execute during periods where the value is within a certain range. The plan's \textit{schedule} corresponds to the chosen time points for the happenings such that all such constraints are satisfied. 

\begin{defn}\label{def:ScheduleDependentFluents}
	A numeric fluent, $v$, is \textit{schedule dependent} at a happening, $j$, within a plan, $\pi$, if executing the plan up to happening $j$ can lead to $v$ taking a range of different values that depend on the chosen schedule for the same plan, $\pi$.
\end{defn}

From Definition \ref{def:ScheduleDependentFluents} it follows that an expression over a schedule-dependent numeric fluent is also \textit{schedule dependent}.

\begin{defn}\label{def:ScheduleDependentCondition}
	A numeric condition, $c = \langle e, \bowtie, k \rangle$, over the set of numeric fluents $\vartheta_c \subseteq \vartheta$, is \textit{schedule dependent} at a happening $j$, within a plan, $\pi$, if and only if $\exists v \in \vartheta_c$, where $v$ is schedule dependent at happening~$j$. 
\end{defn}

\begin{defn}\label{def:ScheduleDependentEffect}%
	A discrete numeric effect, $\langle v, \otimes, e \rangle$, with $e$ being an arithmetic expression over the set of numeric fluents $\vartheta_e \subseteq \vartheta$, is \textit{schedule dependent} at a happening, $j$, within a plan, $\pi$, if and only if $\exists v_e \in \vartheta_e$, where $v_e$ is schedule dependent at happening~$j$. 
\end{defn}

Note that a duration-dependent discrete effect is not necessarily schedule dependent. It could well be that the duration constraints of a durative action resolve to a single value.

If the set of schedule-dependent numeric fluents, $\widetilde{\vartheta} \subseteq \vartheta$ is not empty, the temporal constraints, $C$, need to be combined with the linear constraints representing the trajectory of the schedule-dependent numeric fluents. This can be done with an LP. For each happening $i \in \{0, ..., n\}$, a real-valued LP variable corresponding to each time point $t_i$ is created. For each $v \in \widetilde{\vartheta}$, a pair of real-valued LP variables are created for each happening $i$:

\begin{itemize}
	\item $v_i$, representing the value of $v$ before applying the state transition of happening $i$.
	\item $v'_i$, representing the value of $v$ after applying the state transition of happening $i$. 
\end{itemize}

Equation~\ref{eq:linContEff} shows the constraint for all the concurrent linear continuous effects taking place on $v$ between two consecutive happenings, $i$ and $j$, where $\delta v_{i,j}$ is the constant cumulative rate of change, $\frac{dv}{dt}$, of $v$ for that interval. Equation~\ref{eq:discreteEff} shows the constraint for a discrete update to $v$ at happening $i$, where $\Delta_i$ is the change resulting from applying a discrete effect on $v$ at $t_i$. $\Delta_i$ can itself be duration dependent, and expressed in terms of the time interval between two happenings. These are combined with the STN constraints in $C$ and direct variable assignments, such as those of the initial state, or $:=$ effects.

\noindent
\begin{align}
	v_j - v'_i &= \delta v_{i,j} (t_j - t_i) \label{eq:linContEff} \\
	v'_i - v_i &= \Delta_i \label{eq:discreteEff} 
\end{align}

A schedule-dependent numeric precondition $\langle e, \bowtie, k \rangle$ at a happening $i$ is translated to its equivalent LP constraint by indexing all the schedule-dependent numeric variables in $e$ by $i$. For example, the condition $\langle v, \leq, 0 \rangle$ will be translated to $v_i \leq 0$. Similarly, an invariant condition starting at $i$ and ending at $k$ is translated to LP constraints on $v'_i$, on $v_k$, and on both $v_j$ and $v'_j$ for all happenings $j$, where $i < j < k$. 

If there is no solution to the LP, it means that the state is temporally inconsistent. Otherwise, the maximum and minimum values of each $v$ are also computed in separate runs of the same LP, by changing the objective function to maximise or minimize $v'_n$, where $n$ is the last happening in the partial plan so far. These bounds are stored in $V$ for $s_n$. 

Schedule-dependent numeric goal conditions are also formulated as linear constraints. Since goal conditions must be true at the end of the plan, a goal condition on a schedule-dependent numeric fluent, $v$, for a plan of length $n$, would be translated to its equivalent linear constraint on the LP variable corresponding to $v'_n$. For example, the goal condition $\langle \textit{fuel}, \geq, 10\rangle$, indicating that a vehicle needs to have at least $10$ units of fuel left at the end of the plan, would be encoded as $\mathit{fuel}'_n \geq 10$. Goal constraints are added and solved in a separate LP run from the one used to verify temporal consistency, to check whether the goals defined in the problem have been achieved~\cite{Bajada2015}. 

Finally, if a metric function \cite{Fox2003} is specified for the planning problem, it can be used as the objective function of the last LP run when the goal of the plan has been reached. If the metric function is omitted, the \textit{makespan} (the total duration of the plan) is set as the objective function, by minimising $t_n$, the LP variable representing the time of the last happening of the plan. This guarantees local optimality with respect to the chosen metric and the action sequence of the plan. 

\section{Related Work}

The first category of PDDL2.1 Level 4 \cite{Fox2003} compliant planners that support temporal--numeric interaction with duration-dependent effects are those that impose a restriction that the problem must be reducible to an LP. This means that both the constraints and the trajectories of continuous effects between each discrete state transition must be linear. This approach was used in LPGP~\cite{Long2003},  TM-LPSAT~\cite{Shin2005}, and LP-RPG~\cite{Coles2008}, and more recently by the COLIN~\cite{Coles2012} family of planners, including POPF~\cite{Coles2010a,Coles2011} and OPTIC~\cite{Benton2012}. The same approach was also adopted by Scotty~\cite{Fern2015}, which uses flow tubes~\cite{Li2008} to find plans for actions with control variables. 

The COLIN~\cite{Coles2012} family of planners impose a further restriction on continuous effects of durative actions: that the contribution of any durative action to the rate of change of each numeric fluent, $v$, is constant, so that its cumulative rate of change, $\delta v$, only needs to be updated at the start or end of a durative action that directly affects $v$. This excludes the possibility that a durative action contributes a variable amount to the rate of change of a variable, even if that contribution changes only as a step function at discrete points during the execution of that action. 

\begin{figure}[h]
	\centering
	\includegraphics[width=1.0\linewidth]{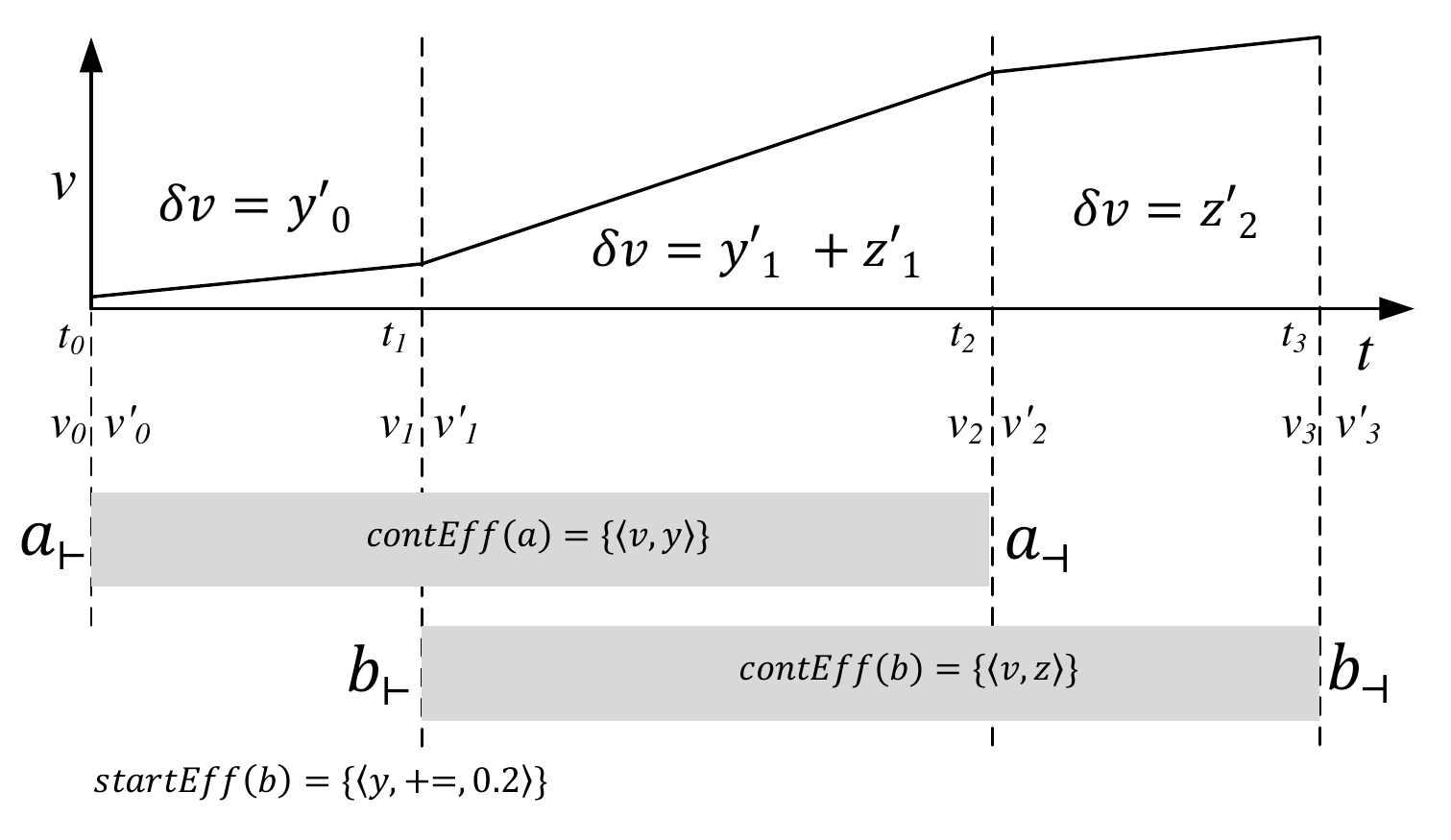}
	\caption[Overlapping Durative Actions]{Overlapping Durative Actions with Interfering Effects.} 
	\label{fig:OverlappingDurativeActions}
\end{figure}

While this approach allows for concurrent durative actions to update the same numeric fluent, it does not allow for updates to the rate of change of a continuous effect itself. The rate of change of a durative action may need to change during its execution due to other interfering discrete effects that update one of the terms of the continuous effect expression. Figure \ref{fig:OverlappingDurativeActions} illustrates an example where durative actions $a$ and $b$ have a continuous effect on $v$, and the snap action $b_\vdash$ has a discrete effect on numeric fluent $y$ which indirectly affects the rate of change of the continuous effect of action $a$.

To support such scenarios, \cite{Bajada2016} propose to account for such discrete updates by recomputing the cumulative rate of change dynamically from the list of actions running concurrently between two consecutive happenings within the plan. This approach removes the restriction imposed by COLIN, POPF and OPTIC, that rates of change of an action have to be constant throughout the plan, and enables full support for piecewise-linear continuous effects. The assumption is that the rate of change is only constant within the context of two consecutive discrete state transitions, thus preserving the linearity of continuous change between adjacent happenings.

LP solvers are quite efficient, and introduce minimal overheads on the typical planning benchmark problem sets that generate small LPs. However, in real-world problems that necessitate long plans (thus larger LPs) and the exploration of thousands of states, LP solving can impact planner performance significantly. Planners typically try to minimise unnecessary LP computation by using only an STN to check for temporal consistency, until an action that has a duration-dependent effect is applied. From then on, the planner uses an LP~\cite{Coles2012}, and no attempt is made to run the LP more selectively and minimise any unnecessary computational overhead.

A profiler-based analysis of the impact of the LP on the planner OPTIC~\cite{Denenberg2019} showed that it spent about 80\% of its time solving the LP, either for consistency checks or to compute the bounds of schedule-dependent numeric fluents. The authors propose to reduce the time spent in LP computation by either not calculating the bounds of all schedule-dependent variables (including omitting the bounds computation step completely), or postponing the LP consistency check to when a candidate goal state is found. While these proposed approaches speed up state evaluation, they introduce trade-offs in terms of temporal consistency and completeness. Omitting the calculation of the bounds of a schedule-dependent numeric fluent, $v$, is problematic because any actions with preconditions on $v$ could be incorrectly deemed inapplicable, leading to loss of completeness. Retaining the LP bounds calculation only for active continuous effects addresses this issue partially, but there is still the possibility that a terminated durative action is forced to change its duration due to temporal constraints added subsequently to the plan, thus constraining further the bounds of the numeric fluents affected by its continuous effects. Solving the LP only on candidate goal states introduces a risk of spending a lot of time exploring a deep search branch that was temporally inconsistent all along and could have been pruned out much earlier in the search. This issue is particularly problematic for planners that use FF-style Enforced Hill Climbing~(EHC)~\cite{Hoffmann2001}, since they do not backtrack beyond the last hill climb and restart a completely new Best-First Search when EHC ends up in a dead end. 

Other planners that do not use Linear Programming have also been proposed. The UPMurphi~\cite{DellaPenna2009,DellaPenna2012} family of planners, including DiNo~\cite{Piotrowski2016a}, use a discretise and validate cycle. The timeline of the plan is discretised into uniform time steps and step functions, to be then validated against the continuous model. ENHSP~\cite{Scala2016} also uses a similar technique, but it supports only the numeric aspect (PDDL2.1 Level 2). The trade-off of this discretization approach is that a significant number of intermediate states are introduced, impairing the ability of these systems to scale. In practice, this approach is only usable for smaller problem instances where non-linearity needs to be preserved. Another alternative is to use SMT. dReach~\cite{Bryce2015} is based on the dReal SMT solver~\cite{Gao2013} and is limited to only a subset of the capabilities supported by the other planners since its focus is mostly on verification of hybrid systems. Furthermore, it uses its own specific language instead of the de facto standard PDDL. ITSAT~\cite{Rankooh2015} encodes the temporal planning task as a Boolean Satisfiability~(SAT) problem, without any numeric support. SMTPlan~\cite{Cashmore2016} encodes PDDL+ into SMT, and supports a suite of non-linear polynomial dynamics. SMT-based approaches also trade off performance, although attempts to optimise the encodings to alleviate this problem have been made~\cite{Rintanen2015,Rintanen2017}. Finally, planners like TFD \cite{Eyerich2009} avoid this problem by restricting numeric effects to occur at the start and the end of an action. 

In this work, we focus on improving the efficiency of planners that use linear programming to handle the interaction between temporal constraints and duration-dependent numeric change. We propose to reduce some of the computational overheads introduced by the LP by using a more selective approach to decide when the LP for the current state needs to be solved, without introducing consistency or completeness trade-offs. We also optimise the LP encoding of the plan to generate smaller LPs that are faster to solve, reducing further the overheads of solving the LP.

\section{Methodology}

To minimise the impact of the LP solver on the planning process, we propose to:

%\begin{enumerate}[label=\alph*)]
\begin{enumerate}
	\item Run the LP solver more selectively, such that it is only executed when the applied action imposes new constraints on schedule-dependent variables.
	\item Reuse predecessor states to minimise the number of LP runs needed to evaluate schedule-dependent goals. 
	\item Optimise the LP encoding, such that it is smaller and faster to solve.
\end{enumerate} 

\subsection{Selective LP Execution}

As described in Section~\ref{sec:Background}, the planner needs to execute the LP for two reasons. Firstly, it needs to confirm whether the plan to reach the state being evaluated can be scheduled consistently. Secondly, it needs to compute the bounds of schedule-dependent numeric fluents, to determine which actions with schedule-dependent conditions are applicable.

If the predecessor of the current state was consistent, the action to reach the current state can make the plan temporally inconsistent if it introduces a negative cycle~\cite{Dechter1991}. If the conditions (preconditions and invariants) of the new action are not schedule dependent, the STN is sufficient to determine consistency, as long as the predecessor state considered all the numeric and temporal constraints when determining its own consistency. Since each temporal state carries an STN of temporal constraints, as described in Section~\ref{sec:Background}, obtaining an updated STN that includes the additional constraints of the new state is more computationally efficient than running an LP on the whole plan~\cite{Muscettola1998,Planken2010,Planken2012,Lee2016}.  

\begin{figure}[h]
	\centering
	\includegraphics[width=1.0\linewidth]{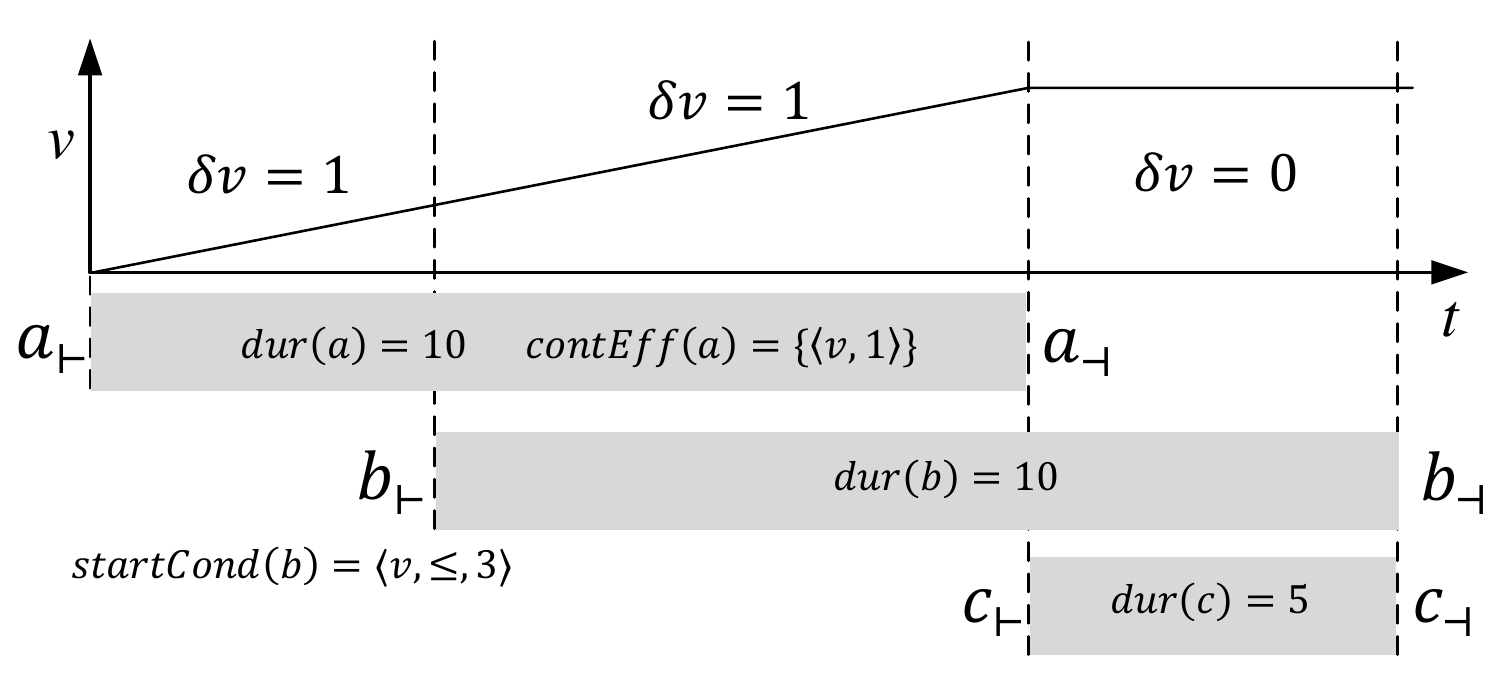}
	\caption[Inconsistent State]{Inconsistent State due to interaction between Temporal and Numeric Constraints, detected by solving an LP but not the corresponding STN.} 
	\label{fig:StnLp}
\end{figure}

A subtle aspect of this process is that if tighter bounds on the temporal constraints were found after computing the LP, they need to be reflected back into the STN, $C$, as a post-scheduling step. The sequence of actions depicted in Figure \ref{fig:StnLp} is one example where this step is crucial. In this case, actions $a$ and $b$ both have a duration of $10$ and overlap each other in their execution. $c$ starts after $a$ ends, and terminates before $b$ ends, with a fixed duration of $5$. The caveat in this example is that  $\stAct{b}$ needs $v <= 3$ to be true. The LP will discover that the latest $b$ can start is 3 time units after $a$ starts. Thus, even though $c$ and $\enAct{b}$ do not refer to any schedule-dependent numeric fluents, the default STN behaviour would incorrectly classify this state as consistent. The upper bound duration between $\stAct{a}$ and $\stAct{b}$ found by the LP needs to be reflected back into the STN of the temporal state obtained after applying $\stAct{b}$ so that it is propagated to the STN of subsequent states.  

For a happening, $j$, whose corresponding action has no schedule-dependent conditions, an STN-based consistency check is enough. Furthermore, if the action has no schedule-dependent effects, the numeric bounds of schedule-dependent numeric fluents computed from the latest prior happening for which the LP was solved, $i$, can be safely carried forward without recomputing the LP. One must keep in mind that these bounds are actually optimistic, because there is a possibility that temporal constraints added after happening $i$ actually restrict the bounds further by making the schedule tighter. If the chosen successor is preconditioned on a schedule-dependent numeric fluent, the LP is then solved for the resultant state, pruning it out if it turns out to be inconsistent due to tighter bounds. 

\subsection{Schedule-Dependent Numeric Goals} \label{sec:DurationDependentGoals}

Apart from participating in action conditions, schedule-dependent numeric fluents can also be part of the goal condition. While the COLIN family of planners \cite{Coles2012,Coles2010a} do not fully support this functionality, it can easily be included in the LP \cite{Bajada2015} as described in Section \ref{sec:Background}. This is achieved by adding the linear constraints corresponding to the goal condition on the values obtained after executing the plan's last happening. 

If the goal condition is schedule dependent, every state discovered during the search process is a candidate for an LP-based goal check, which introduces further computational overhead. To avoid performing this in every state, the information computed for ancestor states can be used to perform the LP-based goal check more selectively. An extra flag is maintained in each temporal state, which primes the LP-based goal check for potential execution. This is turned on when an action has a schedule-dependent effect on a numeric fluent that is featured in the goal condition. The value of this flag is passed on to the successor states until one is found where there are no running actions~\cite{Fox2003} and the goal conditions that are not schedule-dependent are satisfied. If the flag is set to \textit{true}, the LP-based goal check is then executed to check the schedule-dependent goals and the flag is turned back to \textit{false}. This way, the LP-based goal check is only performed if all other conditions that are computationally cheaper to check are satisfied, and we know that since the last LP-based goal check there was at least one update to the schedule-dependent numeric fluents featured in the goal. 

Figure \ref{alg:LpGoalCheck} shows the algorithm described above to check schedule-dependent goal conditions more efficiently. The set $\widetilde{\vartheta}_{s'}$ corresponds to the set of schedule-dependent numeric fluents in state $s'$. The function $\mathit{terms}()$ takes a condition and returns the set of numeric fluents which feature in that condition. The function $\mathit{affected}()$ returns the set of numeric fluents that are affected by an action. The flag that primes the LP goal check is denoted by $lpgc$, which is initialised to $false$ in the initial state. $Q$ corresponds to the list of running durative actions as described in Section \ref{sec:Background}. The function $\mathit{nsd}()$ takes a conjunction of conditions and returns the atomic conditions that are not schedule dependent. The $\mathit{lpGoalCheck}()$ is the function that performs the LP-based goal check on the schedule-dependent fluents of a state, returning $\mathit{true}$ if all are satisfied, and $\mathit{false}$ otherwise. 

If one of the schedule-dependent variables affected by the applied action, $a$, features in the goal condition, the $lpgc$ flag is set to $true$ in the new state, $s'$. Otherwise the same value of the previous state is copied into the new state. If there are no running durative actions and all goal conditions that are not schedule-dependent are satisfied, the algorithm proceeds to check for schedule-dependent goals. If there are none, it can return $\mathit{true}$ immediately since all goals are satisfied. Otherwise, if the $lpgc$ flag is $\mathit{true}$, the LP-based goal check is done and its result is returned. The $lpgc$ flag is also reset to $false$, so that if further successor states are explored the LP-based goal check is not performed again until another action with a schedule-dependent effect on the goal is applied. This algorithm is a drop-in replacement to the standard goal check that is done within the planner's search process, and can be called for every new state being evaluated.

\begin{figure}[h]
	\begin{algorithmic}[1] %[1] enables line numbers
		\REQUIRE The previous state $s$, the applied action $a$, the new state $s'$, and the goal $g$.
		\ENSURE Returns $\mathit{true}$ if $s' \models g$, false otherwise.
		
		\IF {$\exists v \in \widetilde{\vartheta}_{s'} : (v \in \mathit{terms}(g) \land v \in \mathit{affected}(a))$}
		\STATE $s'.\mathit{lpgc} \gets \mathit{true}$
		\ELSE
		\STATE $s'.\mathit{lpgc} \gets s.\mathit{lpgc}$
		\ENDIF
		
		\IF {$s'.Q = \emptyset \land \forall c \in \mathit{nsd}(g) : ( s' \models c)$}
		\IF {$\widetilde{\vartheta}_{s'} \cap \mathit{terms}(g) = \emptyset$}
		\RETURN $\mathit{true}$
		\ELSIF {$s'.\mathit{lpgc}$}						
		\STATE $s'.\mathit{lpgc} \gets \mathit{false}$
		\RETURN $\mathit{lpGoalCheck}(s')$
		\ENDIF
		\ENDIF
		\RETURN $\mathit{false}$
	\end{algorithmic}
	\caption{Algorithm to Check Schedule-Dependent Goals Efficiently.}
	\label{alg:LpGoalCheck}
\end{figure}

\subsection{Optimising the LP Encoding}\label{sec:OptimisingLp}

The LP encoding used in COLIN~\cite{Coles2012} and its family of planners, creates an LP variable, $t_i$, for every happening, $i$, in the plan. This represents the scheduled time of the happening to be resolved by the LP. Furthermore, each schedule-dependent numeric fluent, $v$, is encoded into two variables, $v_i$ and $v'_i$, for each happening, $i$, in the plan as described in Section \ref{sec:Background}. This means that the LP for a plan of $n$ happenings and $m$ schedule-dependent numeric fluents will have at least $n(2m+1)$ variables. While modern solvers do perform some preprocessing steps to eliminate redundant variables and optimise the internal LP structure \cite{Achterberg2020,LPSolve}, we can still take advantage of the knowledge obtained from our plan structure to remove redundant LP variables and constraints more effectively. 

Apart from optimising the LP, removing unnecessary variables and constraints also has other benefits. Encoding a schedule-dependent numeric fluent, $v$, in the LP for every happening of the plan restricts it to linear updates, even for instantaneous effects that take place prior to any schedule-dependent effects on $v$. By encoding the LP variables and constraints only for happenings where the conditions or effects are schedule-dependent, we allow for non-linear instantaneous updates in happenings that do not need to be encoded in the LP at all. In fact, the authors of COLIN~\cite{Coles2012} already bring up the issue of LP efficiency, and suggest adding $v_i$ and $v'_i$ to the LP only when $v$ becomes unstable prior to happening $i$, but the implementation of this family of planners does not take full advantage of this and restricts all effects on $v$ to be linear, irrespective of where they occur in the plan.

From Definition \ref{def:ScheduleDependentFluents} it follows that for $v$ to be schedule dependent at a happening $j$, there should be at least one happening, $i$, where $0 \leq i \leq j$, whose action, $a_i$, affects $v$ in one of the following ways:
\begin{itemize}
	\item $a_i$ has a duration-dependent effect on $v$, and it is a snap action of a durative action that has a non-fixed duration.
	\item $a_i$ is a snap action of a durative action that is still running at happening $j$ and has a continuous effect on $v$. 
	\item $a_i$ has a schedule-dependent effect on $v$.
\end{itemize}

If $i$ is the first happening where $v$ becomes schedule dependent, all prior happenings $h < i$ do not need to have their corresponding $v_h$ and $v'_h$ LP variables. The literal value of $v$ at $i$ will already be known in state $s_i$, and can be encoded directly as a constant from which subsequent LP constraints will be chained in the same way as described in Section \ref{sec:Background}. 

If at any happening $k$, where $i < k$, $v$ resolves to only one value, $v$ is not schedule dependent any more at happening $k$. This includes happenings where $v$ converges to a single value due to tight temporal constraints, or an explicit assignment of $v$ to an expression which is not schedule dependent. Given that $v$ can only take a single value, its corresponding LP variables are not needed any more from happening $k$ onwards (unless $v$ becomes schedule-dependent again). Updates to $v$ at such happenings are thus not restricted to be linear, because they will not be encoded in the LP either. 

Intermediate happenings where the value of a schedule-dependent numeric fluent $v$ is not updated (or its rate of change is 0) can also be eliminated and consolidated into one LP variable. It is sufficient to keep track of the last happening, $i$, that had an effect on $v$ and chain any preconditions or effects to the LP variables of that happening as follows: 

\begin{itemize}
\item A discrete effect on $v$ at a happening $k$, where $i < k$, is encoded as the constraint $v'_k - v'_i = \Delta_k$.
\item A continuous effect on $v$ between happenings $j$ and $k$, where $i < j < k$, is encoded as the constraint $v_k - v'_i = \delta v_{j,k}(t_k - t_j)$. 
\item Preconditions or invariants on $v$ at happening $j$, where $i < j$, are encoded as their equivalent LP constraint with respect to $v'_i$. 
\end{itemize}
 
The LP constraints corresponding to invariant conditions on schedule-dependent numeric fluents can be reduced further by including them at the start happening of the action and only at any subsequent happenings where there is a potentially constraint-violating change. These will be the happenings where the effect updates the constraint's value in the direction of its threshold. For example, if a durative action, $a_{dur}$, has an invariant constraint $v >= 0$, and it starts at happening $i$, the constraint $v'_i >= 0$ is already sufficient for all subsequent happenings during which $a_{dur}$ is executing, as long as $v$ has a non-negative cumulative rate of change, $\delta v \geq 0$, and no discrete effect decreases the value of $v$. 

\section{Evaluation}

The DICE planner \cite{Bajada2016} was enhanced with the proposed techniques to create the optimised version which we refer to as DICE2. In both versions the temporal consistency check is performed after a new unseen state is generated. Additional introspection was added to measure the number of LP runs and performance of both approaches to isolate these gains from the additional planning overhead. The planner's multi-threading capabilities used to evaluate states concurrently were turned off for more accurate measurement of the absolute computation time spent. For each problem, the plan length in terms of happening count (where a durative action results in two happenings), the number of LP runs, and the total time spent solving LPs were recorded.

The total planning time was also compared to that of other popular PDDL2.1~Level~4~\cite{Fox2003} temporal-numeric and hybrid planners, namely POPF~\cite{Coles2011}, OPTIC~\cite{Benton2012}, SMTPlan~\cite{Cashmore2016} and UPMurphi~\cite{DellaPenna2012}. DiNo~\cite{Piotrowski2016a} crashed with a segmentation fault on all problems (this issue was observed by other users and until the time of this writing has not been resolved), and was subsequently excluded. ENHSP~\cite{Scala2016} does not support durative actions and TFD~\cite{Eyerich2009} does not support continuous effects, and thus they were also excluded. Since DICE is developed in Scala and runs on a Java Virtual Machine, and all the other planners are developed in C++, DICE incurs some additional computational overheads mainly from JVM warm-up and non-native runtime inefficiencies. This is more evident in the timings of smaller problem instances. 

The evaluation was done on domains that involve continuous effects and thus require checks that the interaction between the temporal constraints and the schedule-dependent numeric fluents is still consistent in the plan. Experiments were performed on a cloud-based Virtual Private Server with 4 dedicated CPUs, using AMD 2nd Generation EPYC 2.5 Ghz processors with 8 Gb of RAM, running Ubuntu 20.04 LTS 64-bit, using OpenJDK 11.0.7 64-bit. Some of the above planners have deprecated dependencies, which were incompatible with the latest version of Ubuntu. In this case, a Singularity\footnote{https://sylabs.io/singularity/} container was used, using an older version of Ubuntu as the base image together with the required dependencies. This is the same approach that was used in the IPC~2018~competition~\cite{IPC2018,Cenamor2019}. A maximum of 30 minutes was allocated to solve each problem instance.

\subsection{Carpool Domain}

This domain models a cab hailing service where passengers can carpool together if the vehicle has enough capacity. Each problem has several locations, with distance and average speed defined between pairs of locations. Several cars are available, each having fuel and passenger capacity. Trips start at a pick-up location and end at a drop-off location. Trips also include the number of passengers who need to travel together in one car. All the rates of change of the individual continuous effects are constants, and thus all the planners being considered should support this domain. None of the problems involve required concurrency, and all solutions can be sequential. 

\begin{figure}[h]
	\centering
	\includegraphics[width=1.0\linewidth]{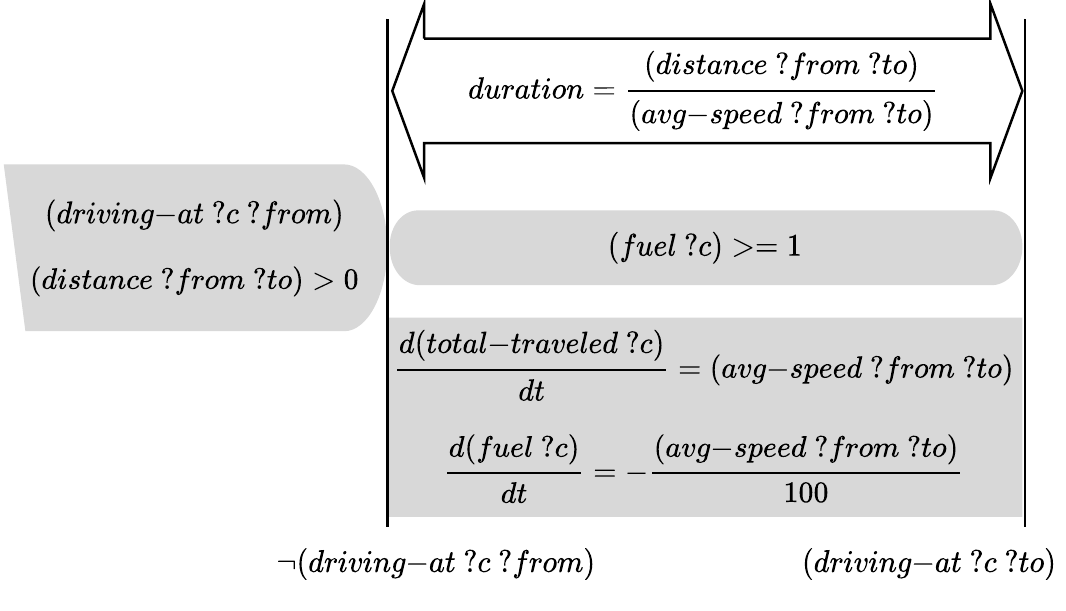}
	\caption[Durative Action]{The \lstinline|drive| Durative Action.} 
	\label{fig:DriveDurativeAction}
\end{figure}

The \lstinline|drive| durative action, depicted in Figure \ref{fig:DriveDurativeAction}, is the only action that has continuous effects in this domain. It keeps track of the total travel distance and fuel consumption, while making sure the car never runs out of fuel:

\begin{lstlisting}[basicstyle=\ttfamily\footnotesize]
(:durative-action drive
:parameters (?c - car ?from ?to - location)
:duration (= ?duration 
  (/ (distance ?from ?to) 
     (avg-speed ?from ?to)))
:condition (and 
  (at start (driving-at ?c ?from))
  (at start (> (distance ?from ?to) 0))
  (over all (>= (fuel ?c) 1)))
:effect (and 
  (at start (not (driving-at ?c ?from)))
  (at end (driving-at ?c ?to))
  (increase (total-traveled ?c) 
     (* #t (avg-speed ?from ?to)))
  (decrease (fuel ?c) 
     (* #t (/ (avg-speed ?from ?to) 100)))))
\end{lstlisting}

The other actions, namely \lstinline|depart|, \lstinline|park|, \lstinline|pickup-trip| and \lstinline|dropoff-trip| are durative actions with only discrete updates that change the state of the car and the trip. The \lstinline|depart| action sets the state of the car to driving, such that embarking passengers is not possible, while the \lstinline|park| action sets the state of the car to stopped, such that passengers can embark or disembark. The duration of both actions is fixed to 1 time unit. The \lstinline|pickup-trip| action embarks the passengers for a trip onto a car, while the \lstinline|dropoff-trip| action disembarks the passengers at their destination. In this case, the duration of both actions is 1 time unit per passenger. A solution for a problem will consist of sequences of the above actions for each trip that needs to be fulfilled. It will typically consist of a \lstinline|depart| action, one or more \lstinline|drive| actions and a \lstinline|park| action with the selected car to reach the trip's pickup location, unless the car is already at that location. This is then followed by \lstinline|pickup-trip|, \lstinline|depart|, and one or more \lstinline|drive| actions to reach the drop-off location. The \lstinline|park| and \lstinline|dropoff-trip| actions executed in sequence then fulfil the trip. While none of the problems feature any required concurrency, having multiple cars available allows for the possibility of concurrent trips served at the same time, if the planner so chooses. 

The problem instances start with 1 trip and 1 car for instance 1, progressing up to 20 trips and 4 cars for instance 20 (a car was added at instances 6, 11 and 16). All problem instances have 100 locations and the goal is to fulfil all trips. 

\begin{table*}[!h]
	\caption{Performance for Car Pool Domain}
	\label{tab:CarPool}
	\centering
	\begin{tabular}{|c|c|c|c|c|c|c|c|c|c|c|}
		\hline
		   &      & \multicolumn{3}{c|}{\textbf{Full LP (DICE)}} & \multicolumn{3}{c|}{\textbf{Optimised LP (DICE2)}} & \multicolumn{3}{c|}{\textbf{ \% Delta} } \\ \hline
		\# & Plan & LP Runs & LP Time (s) & Total Time (s) & LP Runs & LP Time (s) &   Total Time (s)    & LP Runs & LP Time &      Total Time      \\ \hline\hline
		1  &  52  &   54    &    1.863    &     3.254      &   43    &    1.611    &        3.161        &  20\%   &  14\%   &         3\%          \\ \hline
		2  &  64  &   70    &    2.436    &     4.565      &   45    &    1.905    &        3.991        &  36\%   &  22\%   &         13\%         \\ \hline
		3  &  86  &   96    &    5.272    &     7.984      &   57    &    3.596    &        6.267        &  41\%   &  32\%   &         22\%         \\ \hline
		4  &  94  &   107   &    5.815    &     8.855      &   58    &    3.610    &        6.252        &  46\%   &  38\%   &         29\%         \\ \hline
		5  & 168  &   185   &   31.942    &     37.486     &   122   &   24.367    &       29.538        &  34\%   &  24\%   &         21\%         \\ \hline
		6  & 226  &   412   &   99.569    &    114.387     &   242   &   63.445    &       77.351        &  41\%   &  36\%   &         32\%         \\ \hline
		7  & 240  &   553   &   207.513   &    227.337     &   299   &   106.156   &       123.818       &  46\%   &  49\%   &         46\%         \\ \hline
		8  & 248  &   500   &   160.181   &    178.797     &   269   &   82.111    &       98.055        &  46\%   &  49\%   &         45\%         \\ \hline
		9  & 260  &   648   &   235.472   &    259.217     &   337   &   116.417   &       137.198       &  48\%   &  51\%   &         47\%         \\ \hline
		10 & 268  &   599   &   200.275   &    222.179     &   307   &   103.056   &       123.932       &  49\%   &  49\%   &         44\%         \\ \hline
		11 & 312  &  1015   &   437.534   &    468.683     &   443   &   201.972   &       232.181       &  56\%   &  54\%   &         50\%         \\ \hline
		12 & 324  &  1065   &   419.163   &    453.751     &   451   &   215.724   &       247.686       &  58\%   &  49\%   &         45\%         \\ \hline
		13 & 346  &  1187   &   525.192   &    566.956     &   494   &   257.548   &       295.376       &  58\%   &  51\%   &         48\%         \\ \hline
		14 & 354  &  1213   &   529.294   &    571.823     &   496   &   279.023   &       318.202       &  59\%   &  47\%   &         44\%         \\ \hline
		15 & 366  &  1263   &   565.178   &    610.386     &   504   &   275.906   &       315.076       &  60\%   &  51\%   &         48\%         \\ \hline
		16 & 356  &  1239   &   496.731   &    551.682     &   555   &   238.424   &       287.886       &  55\%   &  52\%   &         48\%         \\ \hline
		17 & 360  &  1246   &   466.284   &    518.974     &   542   &   224.392   &       269.826       &  57\%   &  52\%   &         48\%         \\ \hline
		18 & 380  &  1467   &   613.936   &    675.290     &   589   &   248.402   &       303.044       &  60\%   &  60\%   &         55\%         \\ \hline
		19 & 420  &  1783   &   946.034   &   1,024.582    &   688   &   383.345   &       452.479       &  61\%   &  59\%   &         56\%         \\ \hline
		20 & 490  & $\cdot$ &   $\cdot$   &    $\cdot$     &   790   &   706.284   &       800.217       & $\cdot$ & $\cdot$ &       $\cdot$        \\ \hline
	\end{tabular}
\end{table*}

Table \ref{tab:CarPool} shows the performance metrics of both the full LP implementation and the optimised version. All the actions in this domain, apart from the \lstinline|drive| action, do not interfere with the duration dependent variables. Applying these actions does not have any impact on the integrity of the previous LP, and the consistency check of the previous state would still hold for the state obtained after applying one of these actions. When the LP computation was omitted for these states, a significant reduction in planning time was observed. There was a mean improvement of $49.01\% \pm10.73$ in the number of LP runs, and of $44.02\% \pm12.61$ in time needed to compute the LP. The mean improvement on the total planning time was $39.22\% \pm14.52$. The full LP implementation timed out when solving the last problem instance due to exceeding the allocated $30$ minutes, while the optimised implementation solved it in around $13.5$ minutes. As the problems increased in size and complexity, the improvement on planning time also had an upward trend, peaking at $56\%$ for the last instance (19), where both the unoptimised and optimised versions produced a result in time.

This improvement is primarily coming from reuse of memoized information computed in predecessor states when it can be verified that all constraints of the previous LP consistency check have not been disrupted  and no new numeric constraints on schedule-dependent numeric fluents have been introduced. 

\begin{figure}[!ht]
	\centering
	\includegraphics[width=1.0\linewidth]{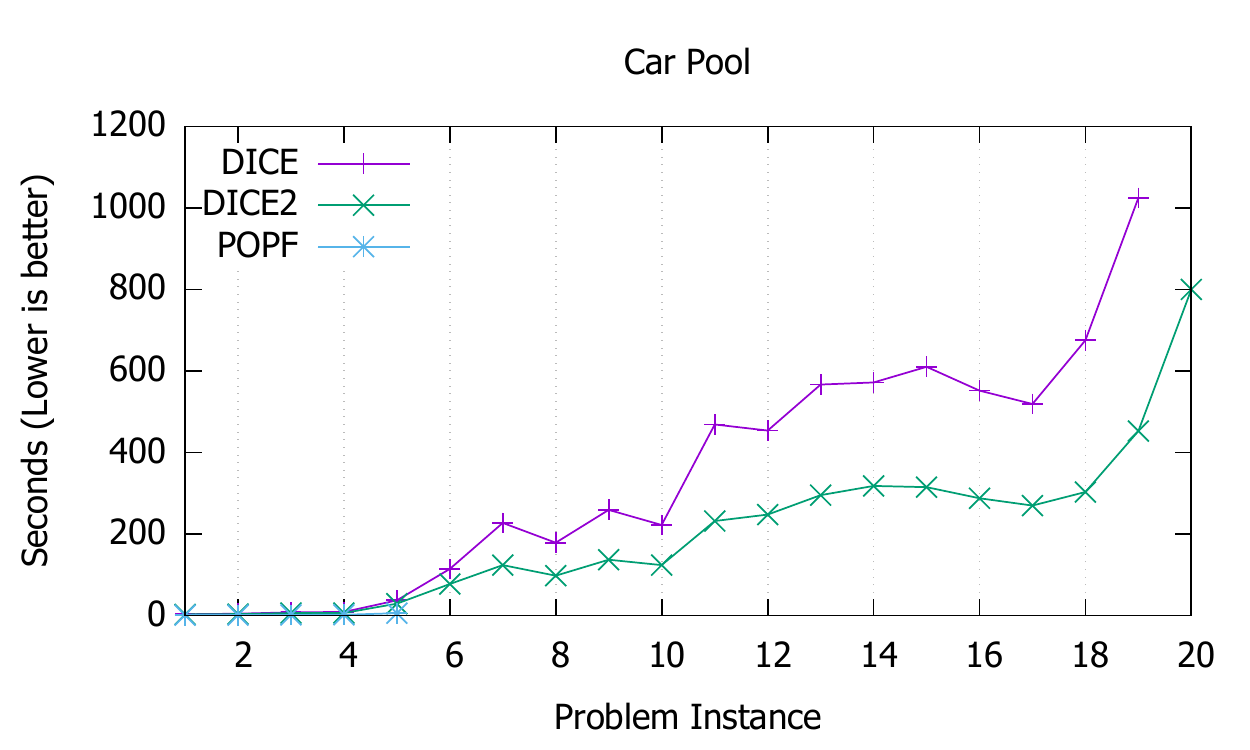}
	\caption[Carpool]{Performance of DICE, DICE2, and POPF on Carpool domain.} 
	\label{fig:Carpool}
\end{figure}

Figure~\ref{fig:Carpool} compares the total planning time of the original and LP-optimised versions of the planner with that of the other PDDL2.1~Level~4~\cite{Fox2003} planners. POPF~\cite{Coles2011} performed well on the first five instances, but timed out due to exceeding the allocated 30 minutes on the rest of the problem instances. OPTIC~\cite{Benton2012} (IPC 2018 version) failed to generate a plan, stopping just after  computing the heuristic of the initial state and starting the search. SMTPlan~\cite{Cashmore2016} also failed with an error on all problem instances. UPMurphi~\cite{DellaPenna2012} was executed with a time-step size of 1 time unit, but it ran out of memory for all problems. Hence, these three planners are not shown.

\subsection{Pump Control Domain}

This domain was originally introduced in \cite{Bajada2016a} and models an industrial plant where various processes need liquid to be pumped at a certain pressure. The objective is to supply these industrial processes with liquid continuously for the duration of their activity without exceeding their pressure thresholds. Tasks also need to be performed, which need some industrial process to be executed either concurrently or prior to the commencement of the task. The planner needs to find the right combination of pump configurations to satisfy the demand without exceeding pressure thresholds. This domain contains piecewise-linear continuous effects, and it is not solvable with the COLIN-style LP formulation~\cite{Coles2012} since the rate of change is not constant. The encoding proposed in \cite{Bajada2016} has to be used.

The \lstinline|fill| durative action increases the \lstinline|current-volume| of the specified resource continuously with the rate computed from the cumulative \lstinline|current-flow-rate| of the pumps. The \lstinline|use| durative action consumes some of the flow for the specified industrial process throughout its duration. The \lstinline|perform-during| and \lstinline|perform-after| actions perform a task during an industrial process (thus introducing required concurrency), or after, respectively. The domain also has the instantaneous actions \lstinline|start-pump|, \lstinline|stop-pump|, \lstinline|increase-pump-flow| and \lstinline|decrease-pump-flow| which start or stop a pump, or increase or decrease its flow respectively. 
A plan consists of \lstinline|start-pump| and \lstinline|increase-pump-flow| actions to bring the pressures to the required levels, followed by the respective \lstinline|fill| or \lstinline|use| durative actions, and \lstinline|perform-during| or \lstinline|perform-after| actions to perform the respective tasks. At the end of the plan, any running pumps must be stopped with  \lstinline|stop-pump|. However, prior to switching a pump off, its flow must be decreased gradually with \lstinline|decrease-pump-flow|.  

The first problem involves 1 pump, 2 industrial processes and 1 task, progressing gradually to the 20th problem with 4 pumps, 16 processes and 25 tasks. 

\begin{table*}[!h]
	\caption{Performance for Pump Control Domain}
	\label{tab:PumpControlResults}
	\centering
	\begin{tabular}{|c|c|c|c|c|c|c|c|c|c|c|}
		\hline
		   &      & \multicolumn{3}{c|}{\textbf{Full LP (DICE)}} & \multicolumn{3}{c|}{\textbf{Optimised LP (DICE2)}} & \multicolumn{3}{c|}{\textbf{ \% Delta} } \\ \hline
		\# & Plan & LP Runs & LP Time (s) & Total Time (s) & LP Runs & LP Time (s) &   Total Time (s)    & LP Runs & LP Time &      Total Time      \\ \hline\hline
		1  &  12  &   58    &    0.145    &     0.833      &   44    &    0.119    &        0.742        &  24\%   &  18\%   &         11\%         \\ \hline
		2  &  18  &   108   &    0.200    &     1.257      &   90    &    0.165    &        1.079        &  17\%   &  18\%   &         14\%         \\ \hline
		3  &  22  &   146   &    0.244    &     1.443      &   126   &    0.227    &        1.454        &  14\%   &   7\%   &         3\%          \\ \hline
		4  &  26  &   182   &    0.303    &     1.499      &   157   &    0.301    &        1.633        &  14\%   &   1\%   &         2\%          \\ \hline
		5  &  30  &   183   &    0.426    &     2.046      &   158   &    0.371    &        1.878        &  14\%   &  13\%   &         8\%          \\ \hline
		6  &  32  &   201   &    0.347    &     2.572      &   74    &    0.170    &        2.005        &  63\%   &  51\%   &         22\%         \\ \hline
		7  &  34  &   248   &    0.428    &     3.009      &   98    &    0.185    &        2.291        &  60\%   &  57\%   &         24\%         \\ \hline
		8  &  38  &   362   &    0.586    &     3.762      &   183   &    0.324    &        2.931        &  49\%   &  45\%   &         22\%         \\ \hline
		9  &  42  &   525   &    0.758    &     4.208      &   338   &    0.624    &        4.108        &  36\%   &  18\%   &         2\%          \\ \hline
		10 &  54  &   538   &    1.689    &     5.660      &   325   &    1.116    &        4.551        &  40\%   &  34\%   &         20\%         \\ \hline
		11 &  56  &   521   &    1.697    &     5.628      &   286   &    1.229    &        5.143        &  45\%   &  28\%   &         9\%          \\ \hline
		12 &  60  &   763   &    2.878    &     8.392      &   493   &    2.228    &        6.897        &  35\%   &  23\%   &         18\%         \\ \hline
		13 &  62  &   777   &    3.047    &     8.742      &   493   &    2.226    &        7.446        &  37\%   &  27\%   &         15\%         \\ \hline
		14 &  66  &   823   &    3.402    &     9.630      &   484   &    1.694    &        6.402        &  41\%   &  50\%   &         34\%         \\ \hline
		15 &  68  &   897   &    3.848    &     10.544     &   524   &    2.196    &        7.732        &  42\%   &  43\%   &         27\%         \\ \hline
		16 &  70  &   987   &    3.426    &     9.670      &   612   &    2.158    &        7.577        &  38\%   &  37\%   &         22\%         \\ \hline
		17 &  72  &  1094   &    5.159    &     12.040     &   734   &    3.619    &        9.758        &  33\%   &  30\%   &         19\%         \\ \hline
		18 &  74  &  1110   &    4.852    &     12.299     &   734   &    2.885    &        8.877        &  34\%   &  41\%   &         28\%         \\ \hline
		19 &  76  &  1214   &    6.289    &     14.026     &   796   &    4.355    &       10.781        &  34\%   &  31\%   &         23\%         \\ \hline
		20 &  86  &  1455   &    7.544    &     17.977     &   943   &    4.240    &       12.892        &  35\%   &  44\%   &         28\%         \\ \hline
	\end{tabular}
\end{table*}

Table \ref{tab:PumpControlResults} compares the LP performance for the full LP implementation and the optimised version. There was a mean improvement of $35.22\% \pm13.58$ in the number of LP runs, and of $30.61\% \pm15.03$ in time needed to compute the LP, across all instances. In this case, the planner could take advantage of state transitions that were not invalidating the prior information about schedule-dependent variables, and thus running the LP again would not yield any additional information. The mean improvement on the total planning time, including search and heuristic evaluation for each state, together with any other runtime overheads, was $17.48\% \pm8.95$. The improvement was not as large as in the Car Pool domain due to the fact that the actions of this domain are very rich in numeric dynamics, and the states where the planner could reuse predecessor information were more limited. 

\begin{figure}[h]
	\centering
	\includegraphics[width=1.0\linewidth]{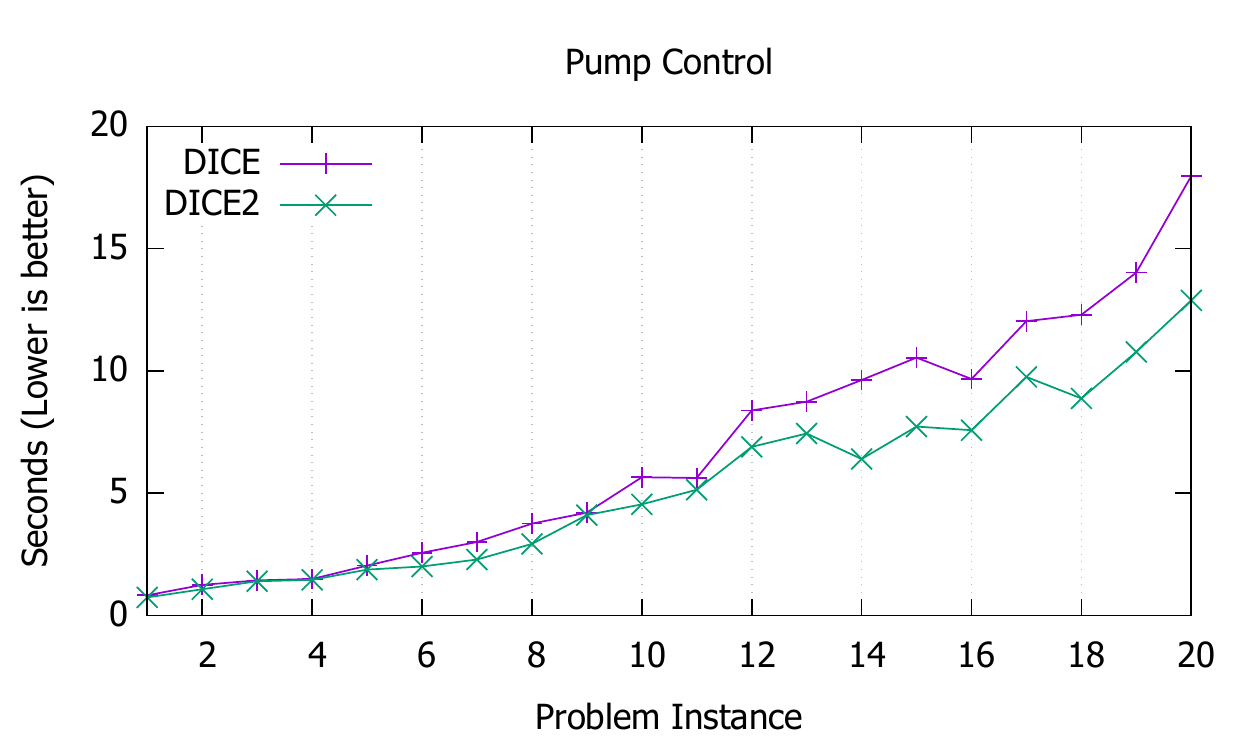}
	\caption[Pump Control]{Performance of DICE and DICE2 on Pump Control domain.} 
	\label{fig:PumpControl}
\end{figure}

Figure \ref{fig:PumpControl} compares the total planning time of the optimised planner with the unoptimised version. In this case, none of the other planners were capable of solving this domain. UPMurphi \cite{DellaPenna2012}, which generates an executable for each problem instance, failed to generate the planner. Both OPTIC \cite{Benton2012} (IPC 2018 version) and POPF \cite{Coles2011} do not support the numeric dynamics of this domain, namely the discrete change to the rate of change of a durative action's continuous effects. 

\subsection{Modified Linear Generator Domain}

This domain is a variant of the Linear Generator domain \cite{Bryce2015,Piotrowski2016a}, which has generators that need to run and tanks which can be used to refuel. It only has two durative actions. The \lstinline|generate| action, which has a fixed duration, switches on the generator and has a continuous decrease effect on the \lstinline|fuelLevel|. The \lstinline|refuel| action has a continuous increase effect on the \lstinline|fuelLevel| of the generator. Once a tank is used to refuel it cannot be used again. A plan consists of just one \lstinline|generate| action and a number of \lstinline|refuel| actions executed concurrently with the \lstinline|generate| action to make sure the \lstinline|fuelLevel| never reaches 0. This domain thus has required concurrency for instances where the generator needs to be refuelled at least once.

The \lstinline|refuel| action was modified to have a variable duration between 8 and 15 time-units, instead of the original fixed duration. This way the planner has to choose how long to spend refuelling, thus utilising the LP to find the solution. An additional schedule-dependent goal was also added to each problem that requires the generator to have at least $10$ units of fuel left at the end of the plan:

\begin{lstlisting}[basicstyle=\ttfamily\footnotesize]
  (:goal (and (generator-ran)
              (>= (fuelLevel gen) 10)))
\end{lstlisting}

\begin{table*}[!h]
	\caption{Performance for Modified Linear Generator Domain}
	\label{tab:LinearGenerator}
	\centering
	\begin{tabular}{|c|c|c|c|c|c|c|c|c|c|c|}
		\hline
		   &      & \multicolumn{3}{c|}{\textbf{Full LP (DICE)}} & \multicolumn{3}{c|}{\textbf{Optimised LP (DICE2)}} & \multicolumn{3}{c|}{\textbf{ \% Delta} } \\ \hline
		\# & Plan & LP Runs & LP Time (s) & Total Time (s) & LP Runs & LP Time (s) &   Total Time (s)    & LP Runs & LP Time &      Total Time      \\ \hline\hline
		1  &  4   &   43    &    0.113    &     0.618      &   34    &    0.101    &        0.541        &  21\%   &  11\%   &         12\%         \\ \hline
		2  &  6   &   71    &    0.166    &     0.836      &   54    &    0.160    &        0.787        &  24\%   &   4\%   &         6\%          \\ \hline
		3  &  8   &   172   &    0.314    &     1.362      &   132   &    0.269    &        1.266        &  23\%   &  14\%   &         7\%          \\ \hline
		4  &  8   &   356   &    0.330    &     2.006      &   272   &    0.322    &        1.904        &  24\%   &   2\%   &         5\%          \\ \hline
		5  &  10  &   713   &    0.346    &     2.858      &   544   &    0.340    &        2.788        &  24\%   &   2\%   &         2\%          \\ \hline
		6  &  12  &  1366   &    0.480    &     4.476      &  1,042  &    0.470    &        4.057        &  24\%   &   2\%   &         9\%          \\ \hline
		7  &  12  &  2334   &    0.519    &     5.622      &  1,776  &    0.501    &        4.967        &  24\%   &   3\%   &         12\%         \\ \hline
		8  &  14  &  3913   &    0.969    &     8.416      &  2,974  &    0.702    &        7.701        &  24\%   &  28\%   &         8\%          \\ \hline
		9  &  16  &  6216   &    1.498    &     11.051     &  4,720  &    0.800    &        9.889        &  24\%   &  47\%   &         11\%         \\ \hline
		10 &  16  &  9246   &    1.178    &     13.311     &  7,011  &    0.945    &       12.300        &  24\%   &  20\%   &         8\%          \\ \hline
		11 &  18  & 13,563  &    2.630    &     20.841     & 10,276  &    1.887    &       17.537        &  24\%   &  28\%   &         16\%         \\ \hline
		12 &  20  & 19,262  &    4.483    &     29.678     & 14,584  &    3.379    &       24.888        &  24\%   &  25\%   &         16\%         \\ \hline
		13 &  20  & 26,242  &    4.871    &     37.851     & 19,852  &    4.801    &       31.546        &  24\%   &   1\%   &         17\%         \\ \hline
		14 &  22  & 35,475  &   11.920    &     53.555     & 26,822  &    8.503    &       45.007        &  24\%   &  29\%   &         16\%         \\ \hline
		15 &  24  & 46,964  &   15.382    &     70.294     & 35,492  &   10.998    &       55.572        &  24\%   &  29\%   &         21\%         \\ \hline
		16 &  24  & 60,976  &   25.242    &     94.320     & 46,060  &   12.813    &       69.927        &  24\%   &  49\%   &         26\%         \\ \hline
		17 &  26  & 77,381  &   24.923    &    110.838     & 58,424  &   20.823    &       93.707        &  24\%   &  16\%   &         15\%         \\ \hline
		18 &  28  & 97,702  &   47.580    &    157.744     & 73,742  &   31.232    &       123.372       &  25\%   &  34\%   &         22\%         \\ \hline
		19 &  28  & 121,696 &   64.020    &    193.642     & 91,822  &   44.749    &       158.695       &  25\%   &  30\%   &         18\%         \\ \hline
		20 &  30  & 148,933 &   83.893    &    245.028     & 112,334 &   54.786    &       190.088       &  25\%   &  35\%   &         22\%         \\ \hline
	\end{tabular}
\end{table*}

Table \ref{tab:LinearGenerator} compares the LP performance between the full LP implementation and the optimised version. In this case the gain was mostly from the optimisations of schedule-dependent goal checks described in Section \ref{sec:DurationDependentGoals}, since both actions in the domain have continuous effects, and the planner didn't find any opportunity to apply selective LP execution or optimise the LP encoding. This resulted in a mean improvement of $23.98\% \pm0.78$ on the number of LP runs and of $20.43\% \pm14.76$ on the time required to compute the LP, across all instances. The mean improvement on the total planning time was of $13.48\% \pm6.29$. 

\begin{figure}[!h]
	\centering
	\includegraphics[width=1.0\linewidth]{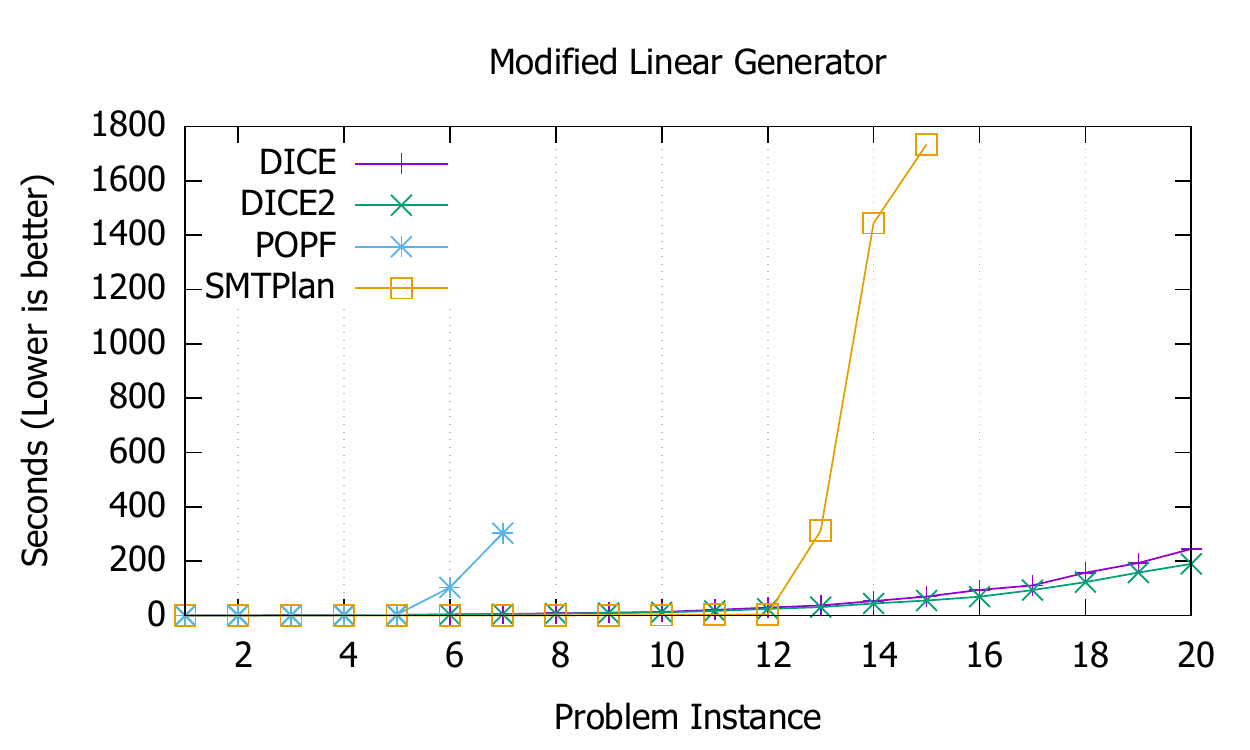}
	\caption[Linear Generator]{Planner Performance on Linear Generator domain.} 
	\label{fig:LinearGenerator}
\end{figure}

Figure \ref{fig:LinearGenerator} compares the total planning time of the original and LP-optimised versions of the planner with that of other PDDL2.1 Level 4 \cite{Fox2003} planners. In this case, UPMurphi \cite{DellaPenna2012}, with a time-step size of 1 time unit, ran out of memory for all problems, and OPTIC \cite{Benton2012} (IPC 2018 version) stopped with an error after estimating the initial heuristic and starting its search. Hence, these two planners are not shown. 

POPF \cite{Coles2011} performed well on the first $5$ instances, slowing down on instance 6 and 7, after which it did not manage to solve any of the problems within the allocated time. All the plans, except for that of instance 3, were actually invalid due to a known bug in POPF when handling schedule-dependent goals. The action sequences of the generated plans are actually valid, but the durations of the actions are not adjusted accordingly to satisfy such goals. For this reason, it was deemed fair to include the results in the comparison just the same. SMTPlan \cite{Cashmore2016} scaled better up to instance 12, after which it also suffered from drastic slowdowns, failing to solve problem instances 16 to 20 within the allocated time. Both DICE and the optimised DICE2 planner scaled more gracefully, with the benefits of optimised schedule-dependent goal checking becoming more visible as the problem complexity increases. 

\section{Conclusion and Future Work}

Schedule-dependent numeric fluents feature in many real-world temporal and hybrid planning problems, and handling them efficiently in a planning system is critical to ensure that a plan is generated within the expected time-frame. 

In this work, we have proposed a number of techniques to improve the efficiency of LP formulation and to only compute the LPs when they are deemed to add new information about the temporal and numeric constraints of the plan. These take advantage of the problem's characteristics especially if it involves actions that do not interact with schedule-dependent numeric fluents. Empirical evaluation using three domains with continuous numeric dynamics showed that these techniques can yield a significant reduction on the number of LP runs and on the actual time needed to compute the LPs, especially if the problem has mixed discrete and continuous actions. This results in a significant overall improvement on the total planning time. Furthermore, the planner outperforms state-of-the-art planners that support the full PDDL2.1~Level~4~\cite{Fox2003} expressiveness in terms of both coverage and scalability, even though it is implemented in a language that has a significantly slower execution runtime.  

The work presented here shows that planners can scale more efficiently to larger real-world planning problems if the information computed during search can be reused across multiple states that have the same constraints rather than being recomputed afresh in each state. Future work includes investigating the possibilities of being more selective on when to run the LP, even for actions that have duration-dependent effects, especially when the bounds of schedule-dependent numeric fluents can be predicted with simpler mechanisms. The support of real-valued control variables and PDDL+ processes can also be investigated and potentially included in the optimised LP encoding. 

\bibliographystyle{IEEEtran}
\bibliography{library}

%%%%%%%%%%%%%%%%%%%%%%%%
\begin{IEEEbiographynophoto}{Dr Josef Bajada}{\space}is a Lecturer in A.I. at the University of Malta. He was previously contracted as an A.I. Technology Consultant with Schlumberger Ltd. to develop A.I. Planning systems for mission critical environments. He obtained his Ph.D. from King's College London, specialising in temporal and numeric planning. His research interests also include intelligent automation, optimisation and reinforcement learning. He also spent over 20 years in an industrial setting, designing and building systems for the Telecommunications and Logistics sectors.
\end{IEEEbiographynophoto}

\begin{IEEEbiographynophoto}{Professor Maria Fox}{\space}is a Principal Researcher in A.I. at British Antarctic Survey, specialising in Planning, Scheduling and Optimisation, and their applications in environmental science. She was previously a Senior A.I. scientist at Schlumberger Ltd, and before that an academic computer scientist and AI specialist. She has published more than 100 papers in A.I. Planning and led a large variety of projects applying planning to problems in energy management, automated drilling and underwater robotics.
\end{IEEEbiographynophoto}

\begin{IEEEbiographynophoto}{Professor Derek Long}{\space}is a Scientific Advisor at Schlumberger Ltd. and Professor of Computer Science at King’s College London. He is a research scientist working in A.I. Planning, particularly focused on temporal and resource-constrained planning. With Professor Maria Fox he developed the widely used standard temporal planning domain description language, PDDL2.1, and extensions of it. He has worked on the application of planning to achieve plan-based automation, in which strategic control of automated systems is determined by a planner. Applications of this work include an automated drilling system currently in active use by Schlumberger Ltd.
\end{IEEEbiographynophoto}

\end{document}